%% file: 0-main.tex
\documentclass[sigconf, authorversion=true]{acmart}

\usepackage{csquotes}
\usepackage{amsmath}
\usepackage{algorithm}
\usepackage[noend]{algpseudocode}
\usepackage{array}
\usepackage{caption}
\usepackage{subcaption}
\usepackage{enumitem}

\usepackage{multirow}
\AtBeginDocument{%
  }

\copyrightyear{2022}
\acmYear{2022}
\setcopyright{acmcopyright}
\acmConference[CIKM '22]{Proceedings of the 31st ACM International Conference on Information and Knowledge Management}{October 17--21, 2022}{Atlanta, GA, USA}
\acmBooktitle{Proceedings of the 31st ACM International Conference on Information and Knowledge Management (CIKM '22), October 17--21, 2022, Atlanta, GA, USA}
\acmPrice{15.00}
\acmDOI{10.1145/3511808.3557067}
\acmISBN{978-1-4503-9236-5/22/10}
\newcommand{\algname}{{e-CLIP}} % Naver Commerce Platform
\newcommand{\Lagr}{\mathcal{L}}

\begin{document}

% \settopmatter{printacmref=false} % Removes citation information below abstract
% \renewcommand\footnotetextcopyrightpermission[1]{} % removes footnote with conference information in first column
\pagestyle{plain}

\title{e-CLIP: Large-Scale Vision-Language Representation Learning in E-commerce }


% \author{Wonyoung Shin$^{1}$, 
% Jonghun Park$^{1}$, 
% Taekang Woo$^{1}$, 
% Yongwoo Cho$^{1}$, 
% Kwangjin Oh$^{1}$, 
% Hwanjun Song$^{2,}$ }
% \authornote{Hwanjun Song is the corresponding author.}
% \email{
%     {wonyoung.shin,  jonghun97.park, t.k.woo, yongwoo.cho, kj.oh, hwanjun.song}@navercorp.com}
% \affiliation{%
%     \institution{$^{1}$NAVER Shopping and $^{2}$NAVER AI Research}
%     \city{}
%     \country{}
%     %\city{Seongnam} 
%     %\country{Republic of Korea}
% }

% \author{Wonyoung Shin$^{1}$, Jonghun Park$^{1}$, Taekang Woo$^{1}$, Yongwoo Cho$^{1}$, Kwangjin Oh$^{1}$, Hwanjun Song$^{2}$}
% \affiliation{%
%     \institution{$^{1}$NAVER Shopping and $^{2}$NAVER AI Research, Seongnam, Republic of Korea \\
%     $\{$wonyoung.shin, ~jonghun97.park,~ t.k.woo,~ yongwoo.cho,~ kj.oh, ~hwanjun.song$\}$@navercorp.com}
% }

% \begin{comment}
\author{Wonyoung Shin}
\email{wonyoung.shin@navercorp.com}
\affiliation{%
  \institution{NAVER Shopping}
  \city{Seongnam}
  \country{Republic of Korea}
%   \postcode{13561}
}

\author{Jonghun Park}
\email{jonghun97.park@navercorp.com}
\affiliation{%
  \institution{NAVER Shopping}
  \city{Seongnam}
  \country{Republic of Korea}
%   \postcode{13561}
}

\author{Taekang Woo}
\email{t.k.woo@navercorp.com}
\affiliation{%
  \institution{NAVER Shopping}
  \city{Seongnam}
  \country{Republic of Korea}
%   \postcode{13561}
}

\author{Yongwoo Cho}
\email{livelyyw93@gmail.com}
\affiliation{%
  \institution{NAVER Shopping}
  \city{Seongnam}
  \country{Republic of Korea}
%   \postcode{13561}
}

\author{Kwangjin Oh}
\email{kj.oh@navercorp.com}
\affiliation{%
  \institution{NAVER Shopping}
  \city{Seongnam}
  \country{Republic of Korea}
%   \postcode{13561}
}

\author{Hwanjun Song}
\email{hwanjun.song@navercorp.com}
\affiliation{%
  \institution{NAVER AI Research}
  \city{Seongnam}
  \country{Republic of Korea}
%   \postcode{13561}
}
\authornote{Hwanjun Song is the corresponding author.}
% \end{comment}



\begin{abstract}
\input{1-abstract}

\end{abstract}

\begin{CCSXML}
<ccs2012>
   <concept>
       <concept_id>10010405.10003550</concept_id>
       <concept_desc>Applied computing~Electronic commerce</concept_desc>
       <concept_significance>500</concept_significance>
       </concept>
 </ccs2012>
\end{CCSXML}

\ccsdesc[500]{Applied computing~Electronic commerce}

\keywords{Multimodal pre-training; Large-scale pre-training}

\maketitle

% \sloppy
% {\fontsize{8pt}{8pt} \selectfont
% \noindent\textbf{ACM Reference Format:}\\
% Wonyoung Shin, Jonghun Park, Taekang Woo, Yongwoo Cho, Kwangjin Oh, Hwanjun Song. 2022. e-CLIP: Large-Scale Vision-Language Representation Learning in
% E-commerce. In \emph{Atlanta '22: The 31st ACM International Conference on Information and Knowledge Management, October 17 - 21, 2022, Atlanta , GA , USA.}  ACM, New York, NY, USA, 11 pages. https://doi.org/XXXXXXX.XXXXXXX
% }

\input{2-introduction}

\input{3-related-works}

\input{4-approach}
\input{5-experiments}

\input{6-conclusion}

% \begin{acks}
% \end{acks}

\bibliographystyle{ACM-Reference-Format}
\bibliography{bibliography}

\end{document}

%% file: 1-abstract.tex
Understanding vision and language representations of product content is vital for search and recommendation applications in e-commerce. As a backbone for online shopping platforms and inspired by the recent success in representation learning research, we propose a contrastive learning framework that aligns language and visual models using unlabeled raw product text and images. We present techniques we used to train large-scale representation learning models and share solutions that address domain-specific challenges. We study the performance using our pre-trained model as backbones for diverse downstream tasks, including category classification, attribute extraction, product matching, product clustering, and adult product recognition. Experimental results show that our proposed method outperforms the baseline in each downstream task regarding both single modality and multiple modalities.

%% file: 2-introduction.tex
\section{Introduction}

% - e-commerce 에 대한 설명
%     - data가 이런식으로 모여서
%     - 모델을 single modality, cross modality task 에 각각 따로 사용하게 되었고.
% - 최근에 clip, align 같은 contrastive learning 메소드들이 나왔다.
% 높은 성능 + large unlabeled data (image + text pair) 때문에 관심을 받게 되었는데
% - e-commerce에 바로 적용하기에는 straightforward 하지 않다. (적용하는게 의미있다.)
% - Challenges
% **: CLIP pre-trained 되어있는게 있는데 바로 가져다 사용하면 좋지 않다. *****
%     - 데이터 측면. (realistic 강조**)
%         - data collection - 수신 받은 데이터를 어떻게 정제하는지 / 이미지 &텍스트 pair 데이터 어떻게 만드는지 (ss_prod service=Y 가 만들어지는데 어려움)
%         - data preparation - 기존 데이터는 정제가 많이 되어있는데 정제하는 과정이 복잡하다 (중복 제거, 텍스트 전처리 등)
%     - Technical 측면 + 엔지니어링적 측면.
%         - Limited resources. 학습이 오래 걸린다.
%         - 데이터가 너무 많아서 로딩하기 어렵다.
% - Novelty
%     - challenge 해결하였다. 해결했으니까,
%     - CLIP method 이  리서치 image task 말고 **realistic한 potential**이 있었다.
%     - CLIP 관련 메소든느 image에만 적용되었지만, text + image 둘다 좋은 성능을 보인다.
%     - 다양한 downstream task에서 성능이 보였고, 다양한 task에도 (0-shot / finetuning)
%     - ~~Single modal, multimodal 체계적으로 분석. 시너지가 나는지.~~
   
\begin{figure*}[t!]
\begin{center}
\includegraphics[width=12.9cm]{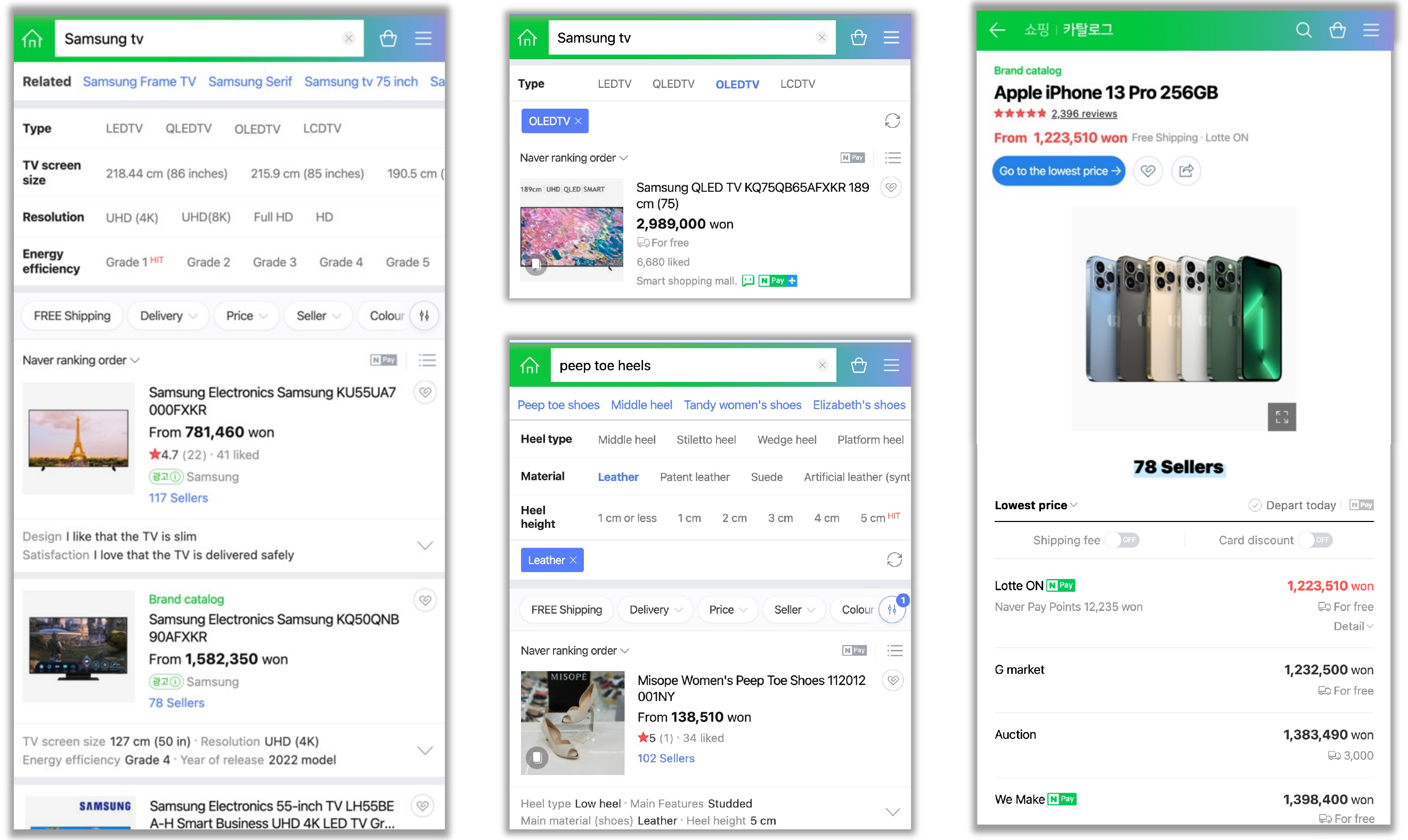}
\end{center}
\vspace*{-0.1cm}
{\small (a) Query-based Search.} \hspace*{1.6cm} {\small (b) Filter-based Search.} \hspace*{1.6cm} {\small (c) Price Comparison.} 
\vspace*{-0.3cm}
\caption{NAVER Shopping is a shopping portal service that provides product search, price comparison, and shopping content so that users can easily access products and sellers registered on NAVER Shopping: (a) shows a search results page with the keyword `samsung tv'. There are various search filters such as categories\,(\textit{i.e.}, type) and attributes\,(\textit{e.g.}, screen size, resolution, and energy efficiency), improving the user's ability to browse relevant products and producing accurate product rankings; (b) shows a search results page after applying category and attribute filters; (c) shows an example of a group of the same products we call `catalog', which helps provide price comparison over multiple sellers on the same product.
\vspace*{-0.3cm}
}
\label{fig:naver_shopping}
\end{figure*} 
    
Product search engines and recommendation systems satisfy millions of users daily and facilitate billion-level transaction records~\cite{sorokina2016amazon, liu2017cascade, zhang2020towards}. As shown in Figure~\ref{fig:naver_shopping}, a diverse set of unimodal and multimodal tasks, including category classification, attribute extraction, product matching, and product clustering, play an integral role in improving the search quality to organize billions of products and enhance user experience systematically~\cite{li2020deepentity, parekh2021fine, hasson2021category}. 

In the industry, training a unified deep neural network jointly across such heterogeneous tasks has been widely applied due to benefits such as reduced operational overhead in managing fewer models, higher accuracy via multi-task learning, and usage as a transferable backbone model for a new downstream task~\cite{beal2022billion, raffel2019t5, bell2020groknet}. However, previous production use cases have mainly focused on training with a single modality~\cite{bell2020groknet, raffel2019t5} or with a framework that is not symmetric for all modalities, thereby restricting the individual use of unimodal models~\cite{lin2021m6}.

%~\cite{pham2021basic, jia2021align}.

Recently, large-scale vision and language pre-training methods, such as CLIP\,\cite{radford2021learning} and ALIGN\,\cite{jia2021align}, have gained considerable research interest in learning a transferable model for multiple downstream tasks. This line of research trains multimodal models jointly based on a contrastive loss such that embedding representations of paired images and texts are similar while those of non-paired images and texts are dissimilar, and these methods exhibit several advantages: (1) They demonstrate promising accuracy on multiple downstream tasks. In some instances, high accuracy can be achieved without fine-tuning on task-specific data~\cite{pham2021basic}. 
(2) These methods comprise two independent models for each modality so that each unimodal model can be used independently for a modality-specific downstream task. (3) These methods can leverage vast amounts of readily available data in an unsupervised manner without labels. 

%, thus potentially eliminating the need for a large amount of labeled training data for a new application~\cite{pham2021basic}; 

%Despite these benefits, the application of these large-scale multimodal pre-training methods has yet been studied or implemented.
Despite their remarkable benefits, the application of these large-scale multimodal pre-training methods has yet to be studied or implemented in the industry. Applying them in the e-commerce industry is nontrivial and poses five challenges concerning the `quality' and `scale' of real-world e-commerce data\footnote{NAVER Shopping has over 1.5 billion products with 500,000 sellers and affiliate malls, such as eBay Korea and Coupang. NAVER Shopping has more than 30 million weekly active users and over 2 billion monthly searches.
}:
% NAVER is a search platform used by more than 40 million monthly users and %The monthly active user of NAVER Shopping accounts for more than half of NAVER usage. 
% the number of monthly searches on NAVER Shopping is over 2 billion.}:

% 제휴몰개수 전체 500K, CPS 105(대형몰), CPC 55K, SS 444K, 
% MAU 20M, 월간 검색건수 2B

% pair mismatch 비율 계산, duplicate 비율
% 카테고리 데이터의 경우, 판매자가 등록한 카테고리의 약 25%~30% 가 틀림. 대부분의 이유는 세분화된 카테고리를 정확하게 등록하지 않기 때문, 예를 들어 닌텐도 조이콘은 electronics > game console > 주변기기 이지만 사용편의를 위해 electronics > game console > nitendo 로 매핑 해둔 사례가 많음. 이외에도 노출 순위를 의도적으로 조작하기 위한 카테고리 변경 행위들이 반영됨

\vspace*{-0.05cm}

\begin{enumerate}[label={\arabic*.}, leftmargin=*]
\item \textbf{Noisy Data.}
Numerous people register products, resulting in noisy data and even product text and image pairs mismatches. The product title can contain irrelevant information about the product, thus hindering the generalization of the model.
%\smallskip
\item \textbf{Duplicate and Similar Products.}
The same product can be registered multiple times by different sellers. 
These products with slightly different attributes, such as their size or volume, constitute a great proportion of data.
%Similar products with slightly different attributes, such as their size or volume, constitute a great proportion of data. 
%These duplicate and similar products are
However, these duplicate products are problematic for contrastive learning, because they are the same but treated as different products during training; the contrastive loss pushes away embeddings of the same product.
%contrastive loss pulls image and text embeddings of the same product pair while pushing away embeddings of pairs from other products~\cite{oord2018representation}. 
%  As shown in Figure~....., these products can disturb the stability of training. 
%\item \textbf{Irregular structures and sparse context.}
%\smallskip
\item \textbf{Irregularity and Sparsity.}
Product text in e-commerce consists of condensed details about products and lacks grammatical structure. They are usually short and have a sparse context~\cite{zheng2018opentag}. For instance, 50\% of product titles in our dataset contain fewer than eleven words. Therefore, capturing the correct semantics from product titles can be difficult.
% 75% 16 words
%\smallskip
\item \textbf{Limited Memory.}
The limited memory of deep learning accelerators such as GPUs and TPUs can act as a bottleneck when training a large model along with a large batch size~\cite{pham2021basic, cui2021zerovl}. Furthermore, due to the large data size, data cannot be fully loaded and can exceed the memory capacity during training.
%\smallskip
\item \textbf{Model Convergence.}
Training with a large batch usually results in lower model accuracy and slow convergence speed, even for a large number of epochs~\cite{song2020large, he2016deep}. %, \textit{e.g.}, 90 epochs in ImageNet-1K training~\cite{he2016deep}. 
Therefore, there remains a need for faster convergence.
\end{enumerate}

% \vspace*{-0.4cm}
In this work, we provide insights from our experience and a series of successfully implemented techniques to overcome the above challenges. 
We first introduce the overall system architecture of NAVER Shopping, and propose a contrastive learning framework named \textbf{\algname{}} that learns visual concepts in the product image using the textual product information. % while taking identical products into consideration. <- maybe the reviewer cant understand at this point
We improve upon this framework by proposing new algorithmic and technical approaches, namely {`catalog-based soft labeling'}, {`category-based negative sampling'}, {`multi-stream accumulation'}, and `batch size scheduler'\,(see Section \ref{pretraining-method} for details). They successfully save memory and expedite model convergence, enabling efficient training with satisfactory performance on multiple e-commerce downstream tasks.
%We further improve upon this framework and propose a hard negative mining method based on weakly-supervised labels which helps the model distinguish similar products.
In addition to the framework, we present effective data preprocessing methods, including cleaning noisy data by removing invalid, duplicate, and inappropriate products.
%We also provide details on saving memory and speeding up model convergence to enable efficient training. 
We conduct extensive experiments on five downstream tasks in NAVER Shopping, \textit{i.e.}, product matching, product clustering, attribute extraction, category classification, and adult product recognition. Experimental results demonstrate the effectiveness of the proposed system and confirm our contributions to the improvement in accuracy and efficiency.

In summary, our main contributions are as follows:
% \vspace*{-0.05cm}

\begin{itemize}[leftmargin=*]
\item To the best of our knowledge, this is the first large-scale industry study investigating a unified multimodal transformer model which is symmetric for both modalities. 
%\smallskip
\item We present an efficient yet effective contrastive learning framework \algname{}, which exploits the large-scale NAVER Shopping dataset regardless of the presence of duplicate products. %We also show the potential of learning representations from similar products using hard negative samples. <- too detail.
%\smallskip
\item We identify five main challenges of adopting a large-scale vision and language pre-training in e-commerce and suggest algorithmic and technical approaches to tackle the issues.
%and achieve better performance as well as faster convergence. 
%\smallskip
\item We conduct extensive offline and online experiments to validate the effectiveness of \algname{} in single and multiple modalities. The results show that the proposed framework boost the performance of both visual and language tasks.  % We additionally propose a novel way of evaluating zero-shot transfer for multimodal input.  <- minor novelty
%\smallskip
\item We apply \algname{} to multilingual downstream classification and clustering tasks in a real industrial scenario and obtain positive feedback, which may benefit further research. 

%We apply contrastive learning methods to multilingual downstream classification and clustering tasks in a real industrial scenario and obtained positive feedback, which may benefit further research. 

\end{itemize}

% \vspace*{-0.5cm}

%% file: 3-related-works.tex
% \newpage

% \vspace*{-0.2cm}
\section{Related Works}
% 3pg (1장)

\subsection{Deep Learning in E-commerce}

%but there is a problem in that it is difficult to make a high-quality algorithm with the existing machine learning technique.
Deep learning has been applied to numerous fields and has achieved great improvement in performance\,\cite{ding2021deep, kamilaris2018deep}. Similarly, in the e-commerce industry, there has been a high demand in applying deep learning methods to real-world downstream tasks, such as category classification\,\cite{hasson2021category}, product attribute extraction\,\cite{parekh2021fine, wang2020learning, donati2019fashion}, product clustering\,\cite{wang2020commerce}, product matching\,\cite{li2020deepentity, tracz-etal-2020-bert}, and product knowledge graph embedding\,\cite{xu2020product, dong2020autoknow}. In particular, matching and discovering the same products play an important role in user experience and price comparison. Product categories and attributes are also core pieces of information that become a major source for search retrieval, ranking, and recommendation systems. However, there has been a challenge in making high-quality product embeddings with existing deep learning-based methods for e-commerce.

With the purpose of reducing maintenance and computational costs, prior work in e-commerce has also focused on adopting a single unified model for various downstream applications. For example, GrokNet~\cite{bell2020groknet} and Shop the Look~\cite{shiau2020shop} use multi-task learning to optimize a single embedding or computer vision model for a diverse set of applications. Likewise,~\cite{beal2022billion} pre-trains a unified backbone model via multi-label classification for visual search systems. These production use cases focus on training with a single modality for visual search systems, while our work investigates pre-training with multiple modalities for both visual and language applications. 

\subsection{Transfer Learning}

Transfer learning with pre-trained weights has been successfully used in natural language processing and computer vision\,\cite{zhang2021hornet, xie2021webke}. The main idea is to borrow knowledge learned from a large corpus into a new downstream task. This learning scheme shows better performance and faster convergence than learning from scratch.

Transfer learning generally proceeds in two stages. 
1) \textit{Pre-training}: training with weakly-labeled or self-supervised loss on large amounts of data. 
2) \textit{Fine-tuning}: supervised learning using labels from the downstream task that is to be applied.

In recent years, researchers have developed a variety of pre-training methods to learn and discover representations from large data. 
In the field of natural language processing, a transformer-based model is commonly trained with a self-supervised objective using masked (or replaced) tokens~\cite{devlin2018bert, liu2019roberta, clark2020electra, brown2020gpt, raffel2019t5, chowdhery2022palm,zhang20ebert}. The computer vision field uses a self-supervised contrastive learning method using image augmentations~\cite{he2020moco, chen2020simclr, grill2020byol, caron2020swav}. Recently, to reduce the spatial inductive bias of CNN, the self-attention structure has been applied to images with a transformer architecture called vision transformers\,(ViT)\,\cite{dosovitskiy2021image}. ViT-based pre-training is known to improve the performance of numerous downstream tasks in computer vision\,\cite{liu2021swin, song2022vidt}. In addition to the standard learning setup, the large-scale pre-trained model works well on few-shot learning and operates robustly in learning with imbalanced data\,\cite{karthik2021learning}.

\begin{figure*}[t!]
\begin{center}
\includegraphics[width=15.2cm]{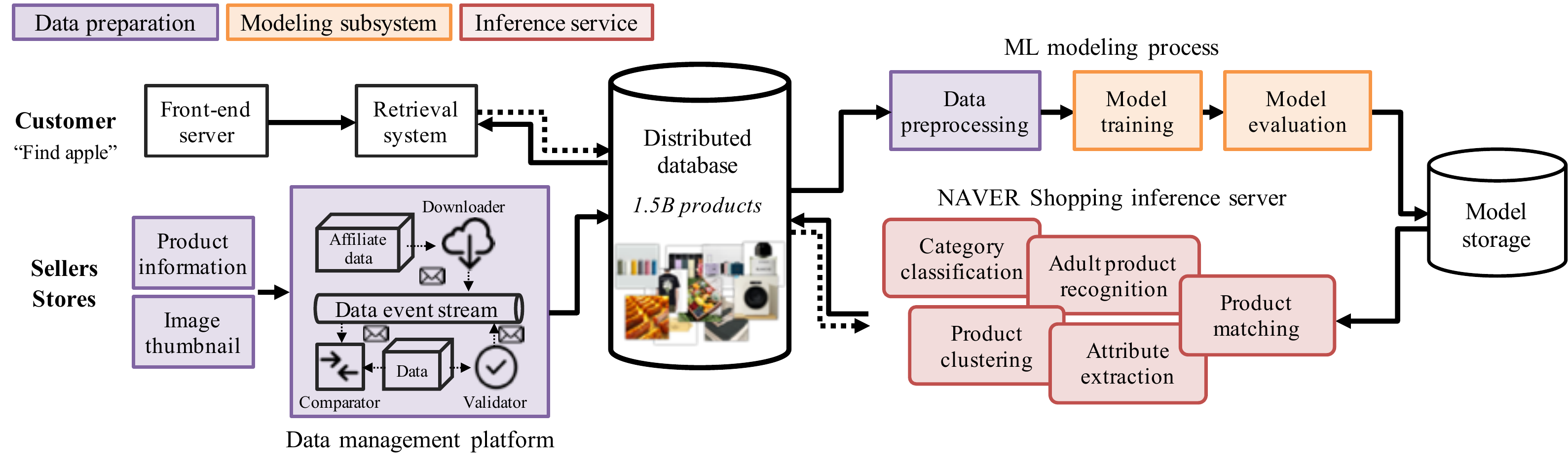}
\end{center}
\vspace*{-0.5cm}
\caption{System architecture of NAVER Shopping. }
\vspace*{-0.4cm}
\label{fig:system_architecture}
\end{figure*}

\begin{comment}
% \begin{figure}[t!]
% \begin{center}
% \includegraphics[width=6.0cm]{figures/invalid.pdf}
% \end{center}
% \caption{\textbf{Removing invalid products }: .....}
% \label{fig:invalid}
% \end{figure} 

% \begin{figure}[t!]
% \begin{center}
% \includegraphics[width=7.0cm]{figures/dup_image.pdf}
% \end{center}
% \caption{\textbf{Removing duplicate images }: .....}
% \label{fig:dup-images}
% \end{figure} 
\end{comment}
    
% \vspace*{-0.3cm}
%\subsection{Multimodality and cross-modality}
\subsection{Multimodal Learning}

E-commerce data comprises rich multimodal data but has not been fully leveraged in the industry. As product data are composed of pairs of product images and text information, multimodal-based representations must be learned for visual and language tasks. 

Multimodal learning employs multiple modalities such as text, image, and audio and has gained much attention, especially after the rise of transformers~\cite{sun2019videobert, tan2019lxmert, sariyildiz2020learning}. Recent works have pre-trained models on large-scale multimodal data to learn generic representations that capture cross-modality relationships~\cite{li2020oscar, cho2021unifying, radford2021learning}. 

A large body of work has trained multimodal models using language modeling and image captioning datasets~\cite{desai2021virtex, chen2020uniter, lin2021m6, tsimpoukelli2021frozen}. Although these methods can learn specific representations for image-text pairs, the framework is not symmetric for all modalities and thus limits the generalization to single-modal scenarios~\cite{yuan2021multimodal}. Many works have also embedded image and text in the homogeneous feature space using their concatenated tokens as input to a transformer~\cite{lu2019vilbert, li2019visualbert, zhang2021vinvl}. Still, learning representations is challenging due to the distribution difference between the visual and text inputs.

As an alternative approach, CLIP\,\cite{radford2021learning} demonstrates the effectiveness of pre-training on a large crawled dataset of approximately 400 million image and text pairs from the internet. CLIP uses contrastive learning by maximizing the similarity between two embeddings generated by image and language encoders for the same item and minimizing the similarity for other items. Several methods have also been proposed to improve performance and resolve scalability challenges. Some researchers presented methods that scale up the model and data size~\cite{jia2021align, pham2021basic}. Some studies add self-supervised learning methods, such as SimCLR~\cite{chen2020simclr}, SimSiam~\cite{chen2021simsiam}, and MLM~\cite{devlin2018bert}, to the existing CLIP loss\,\cite{mu2021slip, li2021declip}. LiT improves the performance of CLIP by locking the weights of the image model and learning only the text representations\,\cite{zhai2021lit}. Meanwhile, ZeroVL presents ways to train CLIP with limited resources~\cite{cui2021zerovl}.

A CLIP-based method can be a potential solution for representation learning due to its simple data preparation and effective training method. However, this family of methods does not provide satisfactory performance in terms of robustness and scalability issues associated with e-commerce data. % \textit{i.e.}, duplicate and noisy data, irregularity and sparsity, heavy operational overhead, and slow model convergence.

%% file: 4-approach.tex
\section{System Architecture}
% 2장.

\input{tab-datafield}

\subsection{Overview}
% - 실제로 돌아가는 end-to-end 학습 파이프라인
% - (data collection -> preparation -> model -> downstream tasks- downstream, 0-shot tasks) 그림 필요.
% - 그림의 step 간단하게 설명

Our methodology involves training and deploying a unified model and powers understanding visual and language representations for search and recommendation systems at NAVER Shopping\,\footnote{https://shopping.naver.com/}. Figure~\ref{fig:system_architecture} illustrates the architecture with three main subsystems: data preparation, modeling, and applications. 

The first stage is data preparation, which is responsible for collecting and managing products daily.
NAVER Shopping receives product content and descriptions, including product titles and images from sellers. The data is then processed and stored systematically in a distributed database.

From the abundant source of data provided by sellers, data is extracted and preprocessed to train our unified multimodal model, as described in~\ref{preprocessing}. Multimodal transformer models are pre-trained on this data using a contrastive loss detailed in Section~\ref{pretraining-method} and saved to storage. If the quality bar is met, the updated model becomes a reliable source model to build upon for downstream modeling tasks, where fine-tuning can be applied to enhance performance. 

Lastly, these downstream applications are used to augment product descriptions with predicted attributes and categories, dramatically impacting the retrieval and ranking of search results. We dive into the details of our framework below, describing each component that participated in significantly improving the performance of applications.

\vspace*{-0.2cm}
\subsection{Data Preparation} \label{data-preparation}
% - 데이터 어떻게 모았는지 + 데이터에 대한 설명 (상품명, 이미지, 브랜드명, 제조사명 등).
% - 모으는데 어떤 어려움이 있었고 어떻게 해결했다. (+쇼핑 수신 팀?)
% - 데이터 관련 통계는 실험쪽에서 보기. Table ### in the experiments section.

\subsubsection{Data Collection} \label{data-collection}
Over a billion products from hundreds of thousands of sellers are deployed, and millions of new products are continuously uploaded, deleted, and modified daily on NAVER Shopping platform. More and more products are registered every day from a growing number of affiliated online shopping malls as well as individual sellers.

According to NAVER Shopping's registration guide, sellers register product content and descriptions, including the product title, price, brand name, and a set of images, as shown in Table \ref{tab:datafield}.
When the registration request is received, product images are stored after processing and resizing to meet the thumbnail image standards of NAVER Shopping. 

% The validity, sold out, and fluctuation of the introduced product are confirmed. Category and catalog information are generated for new products.  NAVER shopping uses product category with up to 4 levels. Product catalog, a bundle of the same products from different sellers, also created to facilitate product comparison. 

The next stage consists of examining whether products are sellable or have been sold out. If products are confirmed to be valid, products are assigned predicted categories through classification and checked if there are similar products in the database for product comparison. Because product information is surprisingly sparse and limited, many parts of NAVER Shopping's product management pipelines are operated with AI-based recommendation and matching systems that use both product images and its textual information (\textit{i.e.}, multimodal input). More details about the prediction tasks are described in Section~\ref{3.4}. Products that have passed verification and classification processes are considered serviceable. Detailed information about the product, such as brand name, category, and product image path, are saved into separate tables and uploaded to the database.

A snapshot of the NAVER Shopping database is generated every day and uploaded to the Hadoop distributed storage engine\,\cite{dittrich2012efficient}. We collect and refine product data from the snapshot data for training. Product data is later used as pairs of a product image and textual information, such as the product title and brand name. To deal with billions of product data, we use Spark~\cite{zaharia2010spark} and mainly work on the Hadoop ecosystem~\cite{laxmi2016analysis}. 
% Refined product data is a pair of a product image and product title text;product title and a brand name.

% \begin{figure}[t!]
% \begin{center}
% \includegraphics[width=7.0cm]{figures/dup_text.pdf}
% \end{center}
% \caption{\textbf{Removing duplicate text }: .....}
% \label{fig:dup-text}
% \end{figure} 

\begin{figure*}[t!]
\begin{center}
\includegraphics[width=15.5cm]{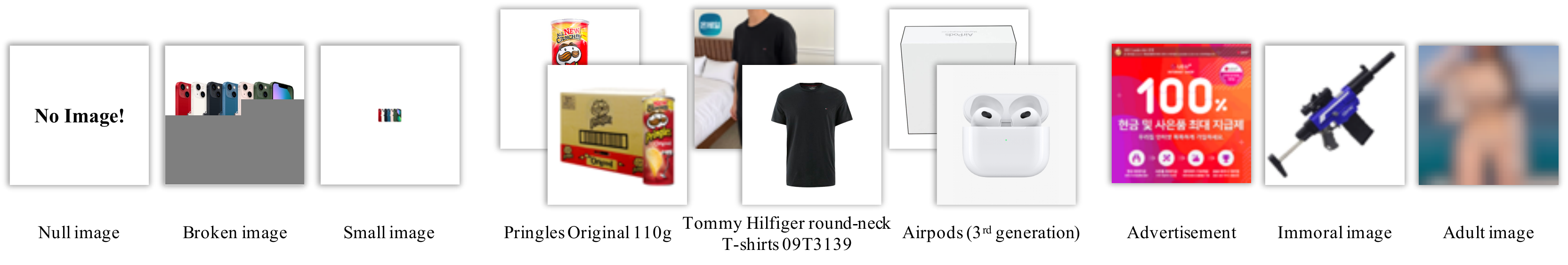}
\end{center}
\vspace*{-0.15cm} %%% \vspace*{-0.15cm}
\hspace*{0.45cm} {\small (a) Invalid products.   } \hspace*{2.7cm} {\small (b) Duplicate products.} \hspace*{2.3cm} {\small (c) Inappropriate products.} 
\vspace*{-0.3cm} %%% \vspace*{-0.3cm}
\caption{\textbf{Examples of filtered products during data preprocessing. }}
\vspace*{-0.3cm} %%% \vspace*{-0.3cm}
\label{fig:data_preparation}
\end{figure*}

%\subsection{Data Preparation}
% \subsubsection{preprocessing} 

% \vspace*{-0.5cm}

\subsubsection{Data Preprocessing} \label{preprocessing}

% 참고: https://oss.NAVERcorp.com/shopping/CLIP/wiki/CLIP-Pretraining-Data
% - 어떤 어려움이 있었고, 어떻게 해결했다. ex) 중복 상품이 정말 많아서 ~~ 방법으로 해결했다.
% - 중복 제거. 텍스트, 이미지 제거.
% - 카탈로그 중복 detection / 제거. (카탈로그 어떻게 만들었고, 수동,자동,라이트 카탈로그 관련 설명 - 매칭 / 클러스터링 알고리즘을 통해서)
% - 텍스트 전처리
% - Train / test split.

% product image and textual information, such as the product title and brand name

% As the product snapshot data is extracted from introduced items, 
% There are a few typical problems found in real e-commerce data;
% abusive information, , mislabeled products,
We use the large volume of collected product data by extracting the product text and image as pairs. We then preprocess the dataset due to common problems that plague e-commerce datasets, including corrupted data, duplicate products, and skewed data distributions. 
% For example, duplicated images decreases the performance of the CLIP-base model as contrastive loss requires diverse examples not redundant per batch. 
We execute preprocessing regarding the following three types of data, which is necessary to improve data quality for better representation learning.
%reduce the problems as much as possible while maintaining the diversity of product data. 
  
\begin{enumerate}[label={\arabic*.}, leftmargin=*]
% \item \textit{Improving Validity of Information:}
\item \textit{Invalid Products}:
Valid pairs of product image and text are essential for cross-modal e-commerce representation learning, but a number of mismatched pairs are included during collection due to system abuse. 
%Valid image-text pairs are essential for cross-modal contrastive learning. However, there are a number of mis-matched pairs with invalid image or text information. 
To improve data validity, we first removed image-text pairs with no images, too small images, and corrupted images, as exampled in Figure \ref{fig:data_preparation}(a), using a rule-based system.  % were also removed using rule-base method.
We also discarded the pairs with invalid text by comparing the token set of product titles; After replacing special characters with white-space in the product titles, we eliminated products with titles consisting of less than two tokens. 
% After white-space tokenized after removing special characters, leaving only English, Korean and numbers. We also eliminated products with titles consisting of less than two tokens. % or only the tokens with extreme low high rarity.
%product names with less than two tokens were eliminated. Product names consist of only tokens with too low or too high rarity have been also excluded. 
%Finally, in order to confirm the validity of product information\,(\emph{i.e.}, the pair of product image and name), the service state of the product was continuously checked.
%\smallskip
\item \textit{Duplicate Products}: % Deduplication
We aimed to eliminate duplicate products, exampled in Figure \ref{fig:data_preparation}(b), based on product titles and images. We first dropped products with identical titles. To eliminate products with duplicate images, we reshaped a product image into (5,5) patches and created one-digit hash keys from the mean color value of each patch. We also took the image size into consideration, resulting in a 29-digit hashed value. Images with the same hashed values were removed. Moreover, as suggested by~\cite{tang2019msuru}, we retrieved compressed embeddings extracted from the last layer of ResNet-34 pre-trained on ImageNet for all images and eliminated duplicate images with the same embeddings.
% 29-digit image hash and model-based image embedding vectors are compared to exclude the same value. We
% Image embedding vector is extracted from a ResNet-34 model pre-trained on ImageNet. 
% We only compare first 256-dimension among 512-dimension of embedding vector for efficiency. 
% The text duplication is dropped based on a white-spaced token sets of product names. 
%% 밑으로 이동!!
% NAVER Shopping creates a catalog of the same products by combining them together through AI recommendation and manual inspection for users to compare products smoothly. We remove products from the same catalog. However, for products grouped in the same catalog, the representative product itself is the same, but different images and different product names can be used. So we leave products belonging the same catalog in the case of n-pair loss learning, and remove with one left from the training data in the other cases. When removing the product, the latest product was left based on the register date of product in snapshot data.
%\smallskip
\item \textit{Inappropriate Products}: %harmful products
Product data collected from various sellers contain inappropriate products that are non-compliant with organizational policies, including adult products, promotional products, social issue products, and immoral products (see Figure \ref{fig:data_preparation}(c)). Not only do these products show distributions that deviate from the general distribution of normal products, but they are also manipulated not to be filtered. For example, sellers blur or mask the critical parts of the image but use product titles that look like normal products. We excluded these products by checking whether the products had been flagged according to the moderation system. 
% and deleted serviced, deleted and the results of NAVER's adult filter inspection. % However, still few amounts of inappropriate products are still included in the training data. Therefore, We include training a CLIP image encoder based sexual product filter in the downstream task.

\end{enumerate}

As a result of preprocessing, our dataset was reduced \textit{from 1B pairs to 330M pairs}, and we call this dataset `NAVER Shopping data.' To shorten training time and compare with baseline methods with the same amount of e-commerce data, we also created a smaller dataset of 270M products. For the latter, we additionally removed products that were predicted to be the same by the product matching system\footnote{We name \algname{} trained on 330M data `\algname{}' while filtered 270M data `\algname{}\,(filter)' throughout the paper.}. The details of the product matching procedure are explained in Section~\ref{3.4}. 
For evaluation during pre-training, 41K products were randomly sampled according to the square root of the natural distribution of product categories and split as test data. 

% After the above preprocessing steps, we finally created a training set containing 270M product data named XXXX. For the n-pair loss model, 330M size training data including catalog overlapped products is generated. 41K products were randomly sampled and split as test data to follow \textcolor{blue}{square root of the actual product category distribution.} (sampled from category distribution of shopping?)

%%%%%%%%%%%%%%%%%%%%%%%%%%%%%%%%%

\begin{figure}[t!]
\begin{center}
\includegraphics[width=8.0cm]{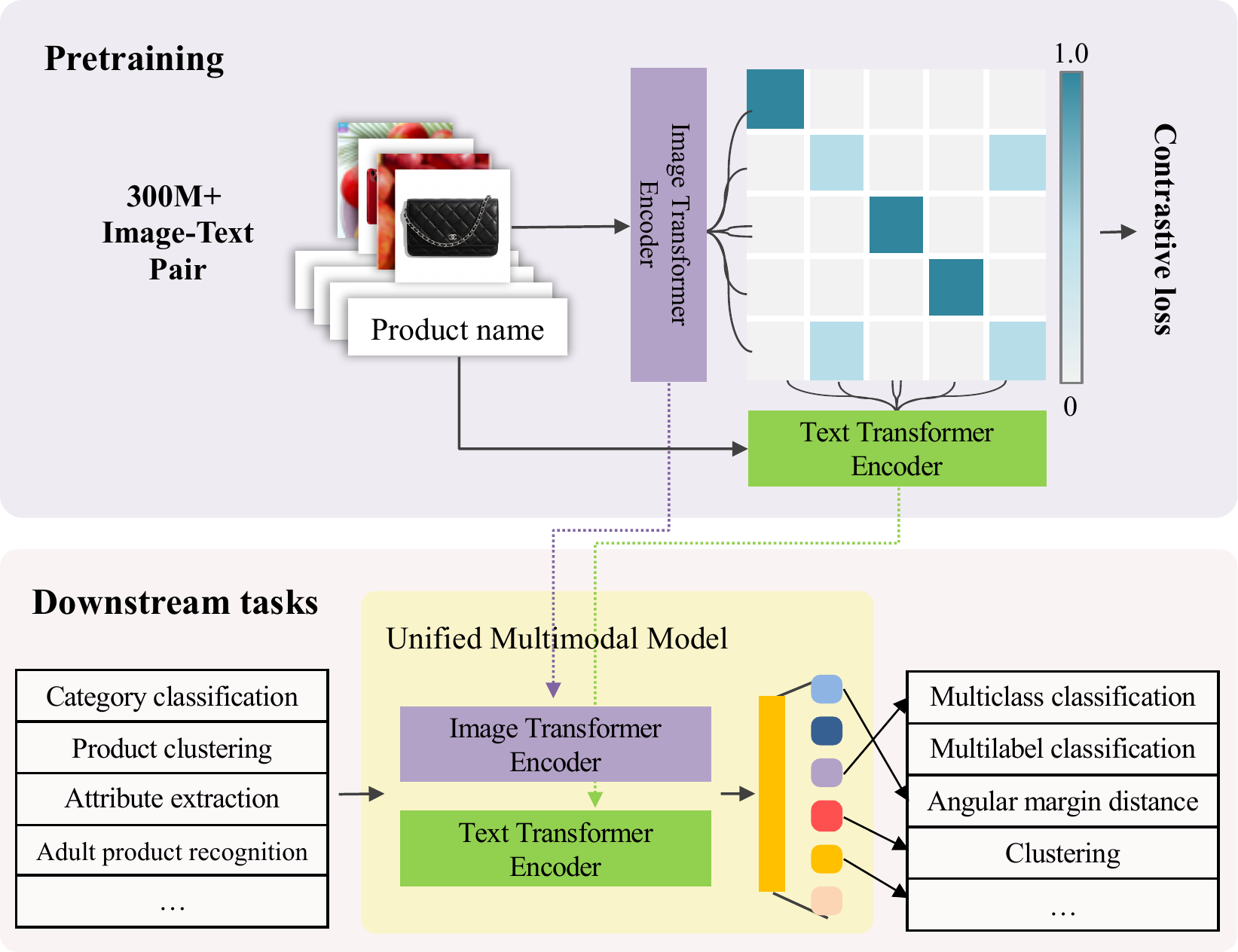}
\end{center}
\vspace*{-0.3cm}
\caption{\textbf{Overview of \algname{} framework. }}
\label{fig:framework}
\vspace*{-0.4cm}
\end{figure}

\subsection{Pre-training Method: e-CLIP} \label{pretraining-method}

Our pretraining-finetuning framework \algname{} is summarized in Figure~\ref{fig:framework}. We first pre-train our multimodal model based on a contrastive loss as suggested in CLIP~\cite{radford2021learning} and LiT~\cite{zhai2021lit}, but consider the presence of duplicate products and optimize the training speed and memory usage of computational resources. Our pre-trained model is later distributed for diverse downstream applications.

% the training schemes

% the ability to embed the various visual concepts in the product to be related to the meta text information the product has. 
% The model is trained under self-supervision using the product image and product text (product name, brand, category, etc.) pairs obtained by the data acquisition method described above (section~\ref{data-collection})

\subsubsection{Model architecture}
% - Transformer 기반 모델 사용. +이유.
% - Pooling layer 변경.

% Visual Genome: Connecting language and vision using crowdsourced dense image annotations
% Inspired by~\cite{jia2021align}, w
% We employ a simple dual-encoder model to align visual and language representations of image-text pairs via a contrastive loss.

Our proposed method \algname{} follows the original dual-encoder structure of CLIP consisting of a visual encoder and text encoder, which return the learned embeddings of the visual and text inputs, respectively. 

% Image and text encoders are ViT-B/32 and multilingual BERT-Base~\cite{devlin2018bert}, respectively. 
For computational efficiency, the image encoder is ViT-B/32, a vision transformer (ViT) mostly composed of multi-head attention blocks~\cite{dosovitskiy2021image, li2021declip}. Although convolutional neural networks (CNNs) are also popular in computer vision tasks, ViTs require shorter training time and fewer GPUs than CNNs in the dual-encoder multimodal pre-training task~\cite{cui2021zerovl,radford2021learning}. To tackle irregularity and sparsity of the data and accelerating model convergence, 
 we fine-tune a pre-trained multilingual BERT-base model~\cite{devlin2018bert} as our text encoder like~\cite{zhai2021lit} and leverage learned weights of multiple languages.
 
%  , and fine-tune pre-trained models for faster convergence.
% Inspired by~\cite{zhai2021lit}, 

Rather than using the representation of the [CLS] token like ALIGN and CLIP~\cite{jia2021align, radford2021learning} or averaging representations of all steps like BASIC~\cite{pham2021basic}, we compute the average of representations across steps of token length at the top layer of our model.

% see a small but consistent improvement by initializing the text model with pre-trained weights

% ko2022large
% CLIP vision transformers are about 3x more compute efficient than CLIP ResNets, which allows us to reach higher overall performance within our compute budget. (CLIP)

% Based on methods such as ... , similar accuracy + more  efficient . LiT showed that contrastive tuning is an efficent way of training with ..

% \vspace*{-0.5cm}
\subsubsection{Training objective}

InfoNCE loss is a popular objective for learning visual models with text supervision~\cite{radford2021learning,pham2021basic}. Image and text encoders are jointly trained using a symmetric cross-entropy loss by maximizing the cosine similarity of paired image and text embeddings while minimizing the similarity of non-paired embeddings in a batch. This contrastive loss expects each sample in the batch to be different from other samples. However, our NAVER Shopping datasets have high volumes of `duplicate' data even after preprocessing. 
Therefore, the same products can become their respective negative pairs, making our contrastive learning pipeline inefficient and ineffective.  

%To address this challenge, wedevelop an objective that effectively learns representations regardless of duplicate products in the batch.
%
To address this challenge, we add a modification to the InfoNCE loss that we call \textit{catalog-based soft labeling}. Let $z_{ij}$ be a label matrix to define whether the image of $i$-th product and the text of $j$-th product are the positive pair. Then, the original CLIP forces only the diagonal entry to be 1, otherwise 0 (\textit{i.e.}, hard labeling). In NAVER Shopping data, the $i$-th and $j$-th product can be the same product, which breaks the assumption of hard labeling. Hence, we internally maintain a list of catalogs, where each catalog is a cluster of the same product in our database and has its unique id. When each product is provided as input to \algname{}, its catalog id is used to find the same products in the batch. Based on the catalog id $c\_id$ , and given a batch of length $N$, we generate a soft label $\tilde{z}_{ij} = 1/|\{j\leq N: c\_id[i]=c\_id[j] \}|$, where $c\_id[i]$ is the catalog id of the $i$-th product and $\sum_{j=1}^{N}\tilde{z}_{ij}=1$ for soft labeling. The catalog and its id are created in a weakly supervised manner since catalog\,(product) clustering and matching are part of downstream tasks (refer to Sections \ref{sec:prod_mat} and \ref{sec:prod_clus}).

% $\tilde{z}

Let ($x$, $y$) be image-text features pairs in a batch of size $N$, and $x_i$ and $y_j$ be image and text features of the $i$-th and $j$-th pair respectively. For the objective of \algname{}, we first compute the cosine similarity between text and image representations as $sim_{i,j}$ for all $N^2$ image-text pairs in a batch.
The softmax function is applied to the similarity matrix, and logits are scaled by a learnable temperature $\tau$ before computing a cross-entropy loss with respect to a uniform distribution over the same product \textit{soft labels} $\tilde{z}_{ij} \in [0,1]$. We formulate a symmetric \algname{} loss as follows:
%Our \algname{} loss uses \textit{soft labels} to ensure that a product can be similar to many products in a batch by leveraging existing data features that store ids of the same products. Our method pulls images and text embeddings of similar products towards each other so that they are mapped to nearby points in the latent space, while different products are mapped to distant points. 

%Let ($x$, $y$) be image-text features pairs in a batch of size $N$, and $x_i$ and $y_j$ be image and text features of the $i$-th and $j$-th pair respectively. We first compute the cosine similarity between text and image representations as $sim_{i,j} = {{x_i^T} {y_j} \over \|{x_i}\| \|{y_j}\|}$ for all $N^2$ image-text pairs in a batch. The softmax function is applied to the similarity matrix and logits are scaled by a learnable temperature $\tau$ before computing a cross-entropy loss with respect to a uniform distribution over the same product labels $z_{ij} \in [0,1]$. We formulate a symmetric \algname{} loss as follows:

% optimize a symmetric cross entropy loss over these similarity scores
% is cosine similarity function computed by dot product of L2 normalized features

%\begin{equation}
%    \ell_{CLIP} = -\frac{1}{2N}(\sum_{j=1}^{N}Y_{j=i}log(I_j\cdot T)+(\sum_{i=1}^{N}Y_{i=j}log(I\cdot T_i))
%\end{equation}
\vspace{-0.2cm} 
\begin{equation}
    \Lagr_{e-CLIP} = \frac{1}{2}\left( \Lagr_{e-CLIP}^{I2T}+\Lagr_{e-CLIP}^{T2I} \right), ~{\rm where}
\end{equation}

% for alignment
{\color{white} empty}
\vspace{-0.5cm} 

\begin{equation}
    \Lagr_{e-CLIP}^{I2T} = -\frac{1}{N}\sum_{i=1}^{N}\sum_{j=1}^{N} \tilde{z}_{ij}  log\left ( \frac{exp(sim_{i,j}) / \tau}{\sum_{l=1}^{N} exp(sim_{i,l})} \right ),
\end{equation}
\begin{equation}
    \Lagr_{e-CLIP}^{T2I} = -\frac{1}{N}\sum_{i=1}^{N}\sum_{j=1}^{N} \tilde{z}_{ij} log\left ( \frac{exp(sim_{i,j}) / \tau}{\sum_{l=1}^{N} exp(sim_{l,j})} \right ).
\end{equation}

Furthermore, it is well known that increasing the difficulty of contrastive learning through hard negative sampling can boost performance~\cite{robinson2020contrastive}. To improve the ability of learning representations, we introduce \textit{category-based negative sampling} based on the product's predicted categories, \textit{e.g.}, ``electronics > wearable device.'' Because we keep each product's category data in our database, as mentioned in Section~\ref{data-collection}, we can easily sample batches composing products with similar category data that have been manually labeled or predicted via classification. Category-based hard negatives help contrast `confusing' products with similar category information; hence, pre-training \algname{} with this technique generally shows superior or comparable performance to its counterparts in the transfer learning setup. 
However, in some zero-shot transfer tasks, it somewhat degrades the performance. Since zero-shot transfer does not have any refinement process like fine-tuning, the pre-trained model with negative sampling can perform poorly because of some noise in the category data despite our efforts for data cleaning.

\subsubsection{Implementation Design} \label{implementation}

% - **Gradient accumulation scheduler** to fasten training
% - FP16, sharding, Gradient checkpointing
% - DGA
% - (잘 안나오면) hard-negative / soft label 고려했다. + ablation test에 추가

%1. Limited memory
%- gpus / tpu memory bottleneck
%- cpu memory bottleneck

%2. Model convergence.

%Limited memory.

% Recently, PLM is known to perform better as the model size increases, so it is becoming common to deal with large models~\cite{}. That much better computing performance is required, but there is a problem that the batch size cannot be increased because the GPU memory is limited compared to the model size. 

% Our aim is to speed up model convergence

Due to larger models and limited memory of computational resources, training with large batch sizes can be infeasible with limited resources. Memory problems mainly occur because of contrastive losses requiring the gradients
of a $NxN$ similarity matrix.
% The main reason memory problems occur is due to contrastive losses requiring the gradients of a $NxN$ similarity matrix, which is memory-intensive. 
Inspired by~\cite{cui2021zerovl,pham2021basic}, we thus implemented a way to train enlarged batches on limited GPU resources through multiple forward passes. 
Our key idea is \textit{multi-stream accumulation},
%The key idea is to 
which computes gradients related to the entire similarity matrix and gradients related to embeddings separately. First, 
embeddings from each forward pass are concatenated and used to calculate the similarity matrix without storing activation values. 
We then execute another stream of forward passes to calculate the gradients for each local batch using the formerly computed similarity matrix and update the model parameters. As a result, by sacrificing reasonable training time, we can successfully train our models with even up to 30x increased batch size.

Speeding up model convergence is challenging due to the large amount of training data, which requires longer training time and limited computational resources. To solve this problem, we develop a convergence algorithm that allows us to complete training models with fewer iterations.

Reducing gradient synchronization frequencies is a common approach in fastening distributed data parallel training~\cite{li2020pytorch,song2020large}. We thus propose a \textit{batch size scheduler} that avoids unnecessary synchronization by replacing the typical learning rate decay scheduler using gradient accumulation. According to \cite{smith2017don}, increasing the batch size produces the same effect as decreasing the learning rate. The learning rate remains constant throughout training and the batch size $B$ for each iteration $t$ is altered as follows:
\begin{equation}
\label{eq:resizing_batch}
B_t \leftarrow B_0 \cdot \left\lfloor 2 \cdot(1+ cos(\frac{\pi \cdot t}{E})^{-1}) \right\rfloor,
\end{equation}
where $B_0$ and $E$ indicate the initial batch size and the number of epochs, respectively. Our batch size scheduling is based on the cosine annealing scheduler without restarts. 

% By accumulating the gradients $\left\lfloor 2 \cdot(1+ cos(\frac{\pi \cdot t}{E})^{-1} \right\rfloor$ times without synchronizing the gradients of parameters, the total communication cost is decreased.

% To overcome the limitation of GPU memory, we apply the gradients accumulation technique to enlarge the batch size. 

% another benefit that gets the total communication cost decreased
% at most to 1/�� of that with constant batch size.

% Model convergence.

% + accumulation scheduler
% We use a strategy of accumulating gradients(GA) for a large enough batch size. Until the accumulation step (over 1) is reached, the gradient is calculated and accumulated as a moving average, and back-propagation is performed with the accumulated grad when the accumulation step is reached. Since gradients are accumulated in train iteration, the learning has the effect of decreasing, and since backprop is performed every accumulation step, the learning has the effect of speeding up.

%%%%%%%%%%%%%%%%%%%%%%%%%%%%%%%%%%%%%%%%%%%%%%%%%%%%%%%%%%%%%%%%%%%%%%%%%%%%%%%%
% multilingual bert, vit setting, batch_size, epoch 같은 하이퍼파라미터도 작성해야할지?

To further speed up training and save additional memory, we use mixed-precision~\cite{micikevicius2017mixed}. The calculation of embedding similarities was also sharded by computing only the subset of the pairwise similarities necessary for each GPU's local batch of embeddings. 

We use the AdamW optimizer~\cite{loshchilov2017decoupled} with a constant learning rate of 3e-5 and schedule the batch size as described in Eq. \eqref{eq:resizing_batch}. The dual encoders of \algname{} are trained on NVIDIA A100 GPUs.
%We use the AdamW optimizer~\cite{loshchilov2017decoupled} with a learning rate initialized to 3e-5 and decayed as described in section~\ref{implementation}. The dual-encoder model is trained on 8 Nvidia A100 GPUs.

%with decoupled weight decay regularization (Loshchilov & Hutter, 2017) applied to all weights that are not gains or biases. 
 % with a batch size of 32,768. 

%%%%%%%%%%%%%%%%%%%%%%%%%%%%%%%%%
%%%%%%%%%%%%%%%%%%%%%%%%%%%%%%%%%

% \vspace{-0.5cm}

\subsection{Downstream Tasks} \label{3.4}
% task 설명 + 그림들

Our pre-trained model is employed in five applications, which are important in e-commerce and has a plethora of research in both academia and industry.

%Our pre-trained model is employed in five applications: product clustering, product matching, category classification, adult product recognition, and attribute retrieval. These tasks are important in e-commerce and each has a plethora of research in both academia and industry.

\smallskip\noindent
\textbf{Product Clustering.} Product clustering denotes grouping the same products and is used to identify new products. A group of products is displayed on a single page with information such as attributes, descriptions, and collected reviews, as shown in Figure~\ref{fig:naver_shopping}c. Discovering new groups from billions of products is complex and requires deep learning models to detect similar products with great precision.

%\subsubsection{Product Matching}
\smallskip\noindent
\textbf{Product Matching.} Product matching refers to the problem of matching products to an existing set of products in the database based on similarity. Identifying whether a product is the same as one from another seller creates the opportunity for users to select the best offer
% after comparing all the available options, 
and prevents users from viewing duplicate products in the search results. Despite its importance, product matching is challenging to tackle, because similar products must be retrieved from a large dataset and product information can vary greatly. 

% . The challenge is further compounded by the fact that product information can vary for the same product. 
% non-trivial task mostly because of the large number of products, their high heterogeneity, missing product representants, and varying levels of data quality

\smallskip\noindent
\textbf{Attribute Extraction.}
%\subsubsection{Attribute Extraction}
Product attributes are defined as the properties of a product. For example, as displayed in Figure~\ref{fig:naver_shopping}b, attributes of a shoe include the material and heel type. Attribute extraction refers to identifying values of an attribute of interest from product data and is used to power faceted retrieval, ranking, and recommendation systems. 
The problem of solving attribute extraction, however, is that acquiring data is difficult; There are multiple attributes for a product, and attributes differ for each product. Attribute values are also often noisy and mostly incomplete with missing values. 

% is an essential component of e-commerce platforms

%\subsubsection{Category Classification}
\smallskip\noindent
\textbf{Category Classification.} E-commerce platforms generally use a predefined structured hierarchy of product categories to enhance user experience and organize products systematically. Accurately matching categories is crucial for improving search quality because categories are used in multiple tasks such as retrieval, ranking, and recommendation.
NAVER Shopping manages over 5,000 categories using a hierarchical category system of up to four levels. 

% For example, women’s shoes can be classified into categories such as ``Fashion goods > Women’s shoes > Flat shoes > Loafers.''
% In e-commerce, product category is basic meta information and is used to limit the product range of applications such as product identification and attribute recognition as well as user's shopping convenience. Categories are hierarchically organized to systematically manage hundreds of millions of products. It consists of 4 levels of “large category > middle category > small category > sub-category”, and for example, women’s shoes can be classified into categories such as “fashion accessories > women’s shoes > shoes > loafers”.

% To improve search accuracy and organize products systematically, the category suitable for the NAVER shopping system is predicted whenever a product is newly registered or updated as reported in section~\ref{data-collection}. 

% Apart from helping buyers sort between different product types, categorization is also critical for multiple downstream tasks, including the platform's main listing search.
%\subsubsection{Adult Product Recognition}
\smallskip\noindent
\textbf{Adult Product Recognition.} Identifying and filtering out adult products is necessary to provide appropriate and safe content for buyers. Because products are uploaded by numerous sellers, some product images can be inappropriate for minors, such as nudity and pornographic images. 
However, it is challenging to develop deep
learning models because not only is collecting flagged products difficult, but the product distributions are heavily skewed; less than 0.01\% of products turn out to be adult products in NAVER Shopping.

% Because products are uploaded by numerous sellers, some products can be unethical or offensive (e.g. ), inappropriate for minors (e.g. ), and unwanted (e.g. advertising).

% The main challenge of adult product detection is the severe imbalance of data; only 0.02\% of products turn out to be adult products in NAVER shopping. 

% NAVER Shopping adult product detection system focus mainly on clothing categories. 
% Although necessary in the field of e-commerce, 

% due to skewed product distributions and varying detection criterion. 

%% file: tab-datafield.tex
{
\newcolumntype{L}[1]{>{\raggedright\let\newline\\\arraybackslash\hspace{0pt}}m{#1}}
\newcolumntype{X}[1]{>{\centering\let\newline\\\arraybackslash\hspace{0pt}}p{#1}}
\begin{table}[!t]
\centering
\resizebox{\linewidth}{!}{%

\begin{tabular}{l|ll}
\toprule
\textbf{Field}    & \textbf{Description}                          & \textbf{Example}   \\
\midrule
Product title      & Title of product & Galaxy watch 4 40mm bluetooth  \\
Brand name        & Brand of product                              & Galaxy                         \\
Maker name        & Maker of product                              & Samsung                        \\
Mall name         & Seller mall                                   & 11 street                      \\
Mall category     & Categories registered by sellers           & Electronics $>$ Watch \\
Price             & Price                                         & 205,000                        \\
Registration time & Date of registration                   & 2022-01-04                     \\
Popularity        & Popularity based on click-through rate        & 32.4                           \\
Image path     & Image URL                     & http://image\_server/image\_id  \\
Product category & Category assigned by NAVER Shopping & Electronics $>$ Wearable device  \\
\bottomrule
\end{tabular}

}
\vspace*{0.00cm}
\caption{Key fields of product information.}
\label{tab:datafield}
\vspace*{-0.8cm}
\end{table}
}

% \begin{tabular}{l|cc|cc|c} \toprule
% \multirow{2}{*}{} & \multicolumn{2}{c|}{Text model} & \multicolumn{2}{c|}{Image model} & \# Pairs \\
%  & Hidden dim & \# Params & Model & \# Params &  \\ \toprule
% BERT-multi & 768 & 167M & - & - & - \\
% BERT-dlarva & 768 & 110M & - & - & - \\
% CLIP & 512 & 37M & ViT/B-32 & 87M & 400M \\
% KELIP & 512 & 88M & ViT/B-32 & 87M & 1.1B \\
% e-LiT & 768 & 167M & ViT/B-32 & 87M & 270M \\
% e-CLIP (small) & 768 & 167M & ViT/B-32 & 87M & 270M \\
% e-CLIP (large) & 768 & 167M & ViT/B-32 & 87M & 330M \\
% \bottomrule

%% file: 5-experiments.tex
\input{tab-model}

% \begin{comment}
% \begin{figure}[t!]
% \begin{center}
% \includegraphics[width=8.0cm]{figures/zeroshot.pdf}
% \end{center}
% \vspace*{-0.3cm}
% \caption{\textbf{Evaluation protocal of zero-shot learning.}}
% \label{fig:zeroshot}
% \vspace*{-0.4cm}
% \end{figure}
% \end{comment}

\section{Experiments}

% 6-9pg (4장)
% \subsection{Pre-training} 

% \subsubsection{Dataset}

%%%%%%%%%%%%%%%%%%%%%%%%%%%%%%%%%

We studied the effect of large-scale pre-training through the transfer performance of \algname{} on the aforementioned five downstream tasks via zero-shot transfer, linear probe, and fine-tuning.

\subsection{Experimental Settings} 

\subsubsection{Comparing Methods}
% We use pretrained vit-b/32 with adam optimizer, cos scheducer, 32,768 batch size and use 37M parameters transformer based text model.% clip text model desc
To validate the effectiveness of our e-CLIP method, we present the results of seven pre-trained models listed in Table~\ref{tab:model}, where ``$\#$ pairs'' indicate the total number of image-text pairs used for pre-training. For a fair comparison, we selected language models pre-trained on a Korean and English corpus. The five baseline models are summarized as follows:
\begin{itemize}[leftmargin=*]
    \item BERT-multi~\cite{devlin2018bert}: Multilingual BERT-base pre-trained on the largest Wikipedia dataset with 102 languages, including Korean and English.
    \item BERT-larva~\cite{kim2019larva}: BERT-base model pre-trained on Korean and English conversation datasets from NAVER services. 
    \item CLIP~\cite{radford2021learning}: Multimodal transformer model which is pre-trained on 400 million image-text pairs and is publicly available.
    \item KELIP~\cite{ko2022large}: Multimodal transformer model pre-trained on 1.1 billion pairs (708 million in Korean and 476 million in English) including 270 million NAVER Shopping products. It should be noted that the product data used to train this model differs from our NAVER Shopping dataset, despite similarities in data size.
    \item e-LiT: Multimodal transformer model pre-trained on our small 270 million NAVER Shopping dataset using our contrastive loss. As proposed by~\cite{zhai2021lit}, we freeze the pre-trained image model weights of CLIP~\cite{radford2021learning} and fine-tune the language model for efficiency and performance.
\end{itemize}

We trained three different versions of \algname{} on our NAVER Shopping dataset with the purpose of (1)\,\textit{e-CLIP vs. e-CLIP (filter)}: examining the effect of rigorous deduplication of products, and (2)\,\textit{e-CLIP vs. e-CLIP (hard)}: investigating the effectiveness of hard negative sampling.
\begin{itemize}[leftmargin=*]
    \item e-CLIP and e-CLIP (filter) are pre-trained on 330 million and 270 million NAVER Shopping datasets, respectively, where duplicated products are filtered out using the AI-based product matching system for the 270 million data.
    \item e-CLIP (hard) is pre-trained on the 330 million NAVER Shopping dataset similar to e-CLIP, but after a warmup stage of training, our category-based negative sampling technique is activated for batch sampling. %to enhance representation learning capabilities.
\end{itemize}

% In this section, we present setting of our pre-training models.
% CLIP learns from 400 million text–image pairs that are already publicly available on the internet. 
% KELIP is a Korean and English bilingual Contrastive Language-Image Pre-training model. Motivated by OpenAI's CLIP. The bilingual multimodal model is trained with collected 1.1 billion image-text pairs , which is three times larger than CLIP's dataset~\cite{}. The text model has 512 hidden dims and 88M parameters, and the image model is Vit-B/32 with 87M parameters.

% LiT, We just teaches a text transformer model to read out good representations from CLIP-pretrained ViT-b/32 image model. % lit desc
% As explained in ~\ref{subsection:pre-training-method}, in order to improve the effect of similar products on contrastive learning, the catalog id of the product was used to reflect the same product. (In other words, it is the id of the same product cluster unit, not the id of the product unit.) If the products in the batch have the same ID after sampling the batch, the corresponding products share the label in equal proportions. For example, if 2 pairs 30 and 55 out of 100 image-text pairs in the batch are identical, {$(I_{30}, T_{30}), (I_{30}, T_{55}), (I_{55}, T_{30}), (I_{55}, T_{55})$} The label of this pair is 0.5.

% Table~\ref{tab:pretrain}

\subsubsection{Evaluation Protocol}
\label{sec:eval_metric}
There are three evaluation schemes, namely zero-shot transfer, linear probe, and fine-tuning.

\smallskip
\noindent\textbf{Zero-shot Transfer.} To measure each method's capability of capturing task-related representations, we evaluate all the methods through zero-shot transfer as proposed by~\cite{radford2021learning}. Prior methods evaluated image encoders using the names of all the classes in the dataset and predicting the most probable (image, text) pair. We extend this method so that both the image and text encoders can be evaluated with multimodal inputs; The image and text representations are averaged to find the most probable class.
% , \textit{i.e.}, the blank token of the given prompt.

\smallskip
\noindent\textbf{Linear Probe and Fine-tuning.} Each task has different evaluation metrics, but the same strategy is used to evaluate learned representations. For linear probe, we freeze the pre-trained model and train a linear classifier to evaluate the embeddings extracted from the model. For fine-tuning, we perform end-to-end training on the entire model, including the encoders and linear classifiers~\cite{radford2021learning}.

\input{tab-catalog-match}

% In the image zero-shot experiment, we confirmed the similarity of image of the target product and the test product image, and so on the text zero-shot experiment. 
% In the multi-modal experiment, an element-wise sum of test image feature and test text feature is used as the test feature. Both target image features and target text features are used as target feature. 

% optimizing the performance of a randomly created proxy.

\input{tab-catalog-cluster}

\subsection{Five Downstream Tasks}
\subsubsection{\textbf{Task 1: Product Matching}}  \label{sec:prod_mat}

Identifying identical products from existing products is highly valued for food products and is challenging for electronic products. We, therefore, validate the capability of recognizing the same product from 300K products from both categories. We average the matching results of almost 200K products for each category.

% We adopt  to evaluate matching performance.
% Experimental results are reported in Table~\ref{tab:catalog-match}. 
As displayed in Table~\ref{tab:catalog-match}, all the \algname{} models generally demonstrate substantial gains in top-1 zero-shot transfer accuracy for both categories. Our best models significantly improve the performance of other BERT-base models by up to 32.6\% and 44.5\% for electronic and food products, indicating that our models are exceptional at learning textual representations. Our \algname{} models also show higher performance on multimodality than other pre-training methods. 
We can also observe that the performance of image experiments are lower, especially for electronic products, due to the task difficulty; For example, ``RTX2080'', ``RTX2080ti'' are different products but their images are almost identical and difficult to distinguish for humans. 
% For this reason, although our methods attain higher multimodal accuracy, using images can, in some cases, cause the accuracy for multimodal inputs to be lower than the text accuracy.

As for our three variations, the results show that catalog-based soft labeling makes \algname{} robust to duplicate products since it generally outperforms \algname{} (filter).
The results of \algname{}\,(hard) show that using a larger training dataset and category-based negative sampling can further enhance the performance of identifying similar products from a considerable number of products.
% the accuracy for multimodal inputs is similar to or slightly improves the accuracy for textual representations.

\input{tab-attribute}

\subsubsection{\textbf{Task 2: Product Clustering}} \label{sec:prod_clus}

% \item \textit{Product Clustering}: Similar to product matching, we evaluate the quality of clustered embeddings on both food and electronic products that have almost 1000 and 2000 number of true clusters, respectively. 

We evaluate the quality of clustered embeddings on both food and electronic products that have almost 1,000 and 2,000 clusters, respectively. Each target cluster contains 20 to 50 products. We evaluate product clustering tasks by extracting embeddings from frozen pre-trained models without additional training.

To assess product clustering, we obtain the multimodal embedding vector by concatenating the text and image embedding vectors and reducing the dimension to 128 using the PCA algorithm~\cite{minka2020automatic,tipping2006mixure}. The projected vectors are later clustered with the $k$-means algorithm~\cite{macqueen67somemethods}. We assessed the clusters using common metrics such as the clustering accuracy (denoted by {ACC}), normalized mutual information (denoted by {NMI})~\cite{strehl2003cluster}, and adjusted Rand index\,(ARI)~\cite{ronen2022deepdpm,zhang2022oamine}. Results in Table~\ref{tab:catalog-cluster} show that \algname{} outperforms baseline models and \algname{}\,(hard) mostly yields the best performance in all cases, providing compelling evidence that (1)\,our methods are effective in distinguishing, and grouping similar products and (2)\,hard negative sampling can provide an additional performance boost.
% support the the fact that using hard negative samples 
% We first extract the product image and text embedding. The multi-modal embedding is a concatenated vector of text and image embedding vectors. The dimension of each embedding vector is reduced from 512 (1024 for multimodal) to 128 through PCA projection. The projected vectors are clustered with the K-mean algorithm.
% The value of K is determined by the number of target clusters, 987 in this task, and each cluster contains about 20 to 50 products. 

% We use typically assessed clustering metric; clustering Accuracy(ACC), Normalized Mutual Information(NMI) and Adjusted Rand Index(ARI). Our result in Table~\ref{tab:catalog-cluster} suggest that \algname{}s outperforms baseline models and \algname{} (large+hard) obtains the best performance on all experiments. 

\input{tab-category-zero}
\input{tab-category-linear}

\subsubsection{\textbf{Task 3: Attribute Extraction}}
% zero-shot attribute retrieval

To evaluate a method's capability of capturing attributes from images, we used a dataset containing 3000 shoe products with five attributes: colors, main materials of the shoe, heel types, toe shapes, and patterns. We evaluate attribute extraction by creating Korean and English prompted attribute value texts as target classes for each attribute. 

Table~\ref{tab:attribute} summarizes the zero-shot transfer results. Inspection of the results indicates that most methods outperform CLIP, which was not successful in understanding concepts specific to the e-commerce industry. Although \algname{}\,(filter) was trained on the same number of e-commerce product pairs as KELIP, e-CLIP\,(filter) achieves higher accuracy than KELIP in most cases. Interestingly, our \algname{} models are effective in predicting attributes such as materials, heel types, and toe shapes, which require capturing fine-grained features of a product and domain knowledge. These results imply that pre-training on domain-specific pairs is more effective in learning target concepts and understanding semantics. 
%relatively small number

% Attributes such as colors and patterns are easy to capture

% KELIP provides comparable results in some attributes. Our methods acts a strong baseline 

% In the ``colors'' and ``patterns'' attribute types, most models including CLIP performs good. For attribute types that need to capture fine-grained features of a product such as ``materials'', ``heel types'', and ``toe shapes'', methods that were trained on our dataset yields higher performance than KELIP. Table~\ref{tab:attribute} summarized this result.

 % The attribute value text feature with the highest similarity compared to the product image feature is selected as the predicted value. 

\vspace*{-0.2cm}
\subsubsection{\textbf{Task 4: Category Classification}}
% zero-shot transfer category
% image, text, multimodal accuracy 뽑는 상세한 방법 설명은 위 쪽에서 나올 것 같음.

To verify the models' performance on categories that are vital to NAVER Shopping, we evaluated models on three datasets: 10K samples of all the categories (denoted by All), 100K samples of fashion goods, and 50K children related products. For linear probe and fine-tuning, we trained models on training data of approximately 21 million products. Because categories are naturally imbalanced, the number of products in each category in the training data distribution was normalized to prevent extreme bias in popular categories. %We use classification accuracy as the evaluation metric. 

% To include additional data, 
% Based on the product title and brand name, if there is arbitrary category name registered by seller, it is added and used as test product text. We also ensemble 12 test product text prompts and 8 target category text prompts. 
Table~\ref{tab:category-zero} reports the test accuracy results of zero-shot transfer. The results show that our models are more proficient at classifying the category of products than other baseline methods. With the same amount of e-commerce training data samples, our methods outperform KELIP by 1.3\% to 30.1\%. The baseline CLIP model scores less than 1\%, presumably due to a lack of Korean training data and data related to the e-commerce domain. Unlike KELIP, our \algname{} methods also achieve consistent and remarkable improvements in accuracy regarding multiple modalities compared to single modalities. These results could be interpreted that not only are representations for single modalities learned competently, but the text and image models are also better aligned in the multimodal feature space than other methods. 

Table~\ref{tab:category-linear} displays the classification accuracy results for linear probe and fine-tuning. Our \algname{} models outperform other baseline methods, including BERT-base models, CLIP, and KELIP, for both linear probe and fine-tuning experiments. It is also noteworthy that there is not a significant difference in accuracy for linear probe and fine-tuning, which could be attributed to the large number of samples in the training data for category classification.

% \algname{} slightly better than KELIP, from XX\% in average in the image linear probe experiment and XX\% in average in the text linear probe experiment. Result of finetuning experiments also have the same tendency. It is also noticeable that the worst \algname{}s' linear probe score beats the finetuned baseline models in any experiments.

% \algname{} models not only learns uni-modal representations well, but also aligns the latent multi-modal feature space better than CLIP and KELIP in the e-commerce domain. 

% \subsubsection{Zero-shot transfer}
% As the target category text is a hard label not considering the hierarchical architecture, the performance of the \algname{} (small) is better than \algname{} (large), \algname{} (large+hard) in most cases. However, \algname{} (large), \algname{} (large+hard) still show similar or better performance than KELIP in most cases.

%  \algname{} model outperforms KELIP in all cases 

% \subsubsection{Representation learning}
\input{tab-adult-linear}

\subsubsection{\textbf{Task 5: Adult Product Recognition}}
We evaluate on adult products flagged by the NAVER Shopping monitoring system. We extract products collected for a year, sample non-adult products, and generate training data consisting of 10K adult products and 50K normal products. The test dataset consists of approximately 5000 normal products and 1000 adult product images. We use F1-scores for evaluation.

% zero-shot adult filter
% We evaluate on a number of prompts that 
% Wethe presence of adult products using prompts in Korean and English. 
 
%  \subsubsection{Zero-shot transfer}
Table~\ref{tab:adult-linear} displays the performance of zero-shot, linear probe and fine-tuning for adult product recognition. 
% As shown in the table, 
Our \algname{} models outperform baseline models on F1-scores by remarkable margins for the zero-shot transfer task. Our models seem to learn domain-specific knowledge better than CLIP and KELIP because they significantly improve precision without damaging recall or detecting adult products excessively. For example, ``lingerie'' is usually treated as non-adult products in NAVER Shopping, but CLIP and KELIP may have been trained on ``lingerie'' images with more provocative texts, causing these products to be predicted as adult products. In addition, our models also achieve the best results for linear probe and fine-tuning.

\subsection{Online Experiments}
% - 서비스보다 ## 향상되었다.
% - 효율성 측면. 리소스 감소.

We confirmed the effectiveness of \algname{} by running online experiments and integrating it with downstream applications. By leveraging our model embeddings directly and our model to train new models,
we observed improvement in several applications and overall training efficiency. For example, we achieved a 14.11\% relative improvement in F1-scores for adult product recognition. Our proposed models are a good fit because they have high-quality representations 
% using textual supervision 
and are strong in transfer learning, as seen in our offline experiments. 
For category classification, we observed comparable performance by simply training a linear classifier.

% relative improvement in classification accuracy through monitoring the system.

% To evaluate the effectiveness of proposed models in real world, we compare \algname{}s with the model in service in the adult product recognition tasks. Serving model is fine-tuned ReXNet\_1.0~\cite{}, and scores 70.953\% f1-score and 68.117\% recall. on test dataset. Linear probed \algname{}s can beat the serving model because large number of shopping domain data has been trained. In particular, our best model scores 78.392\% in f1-score and 80.551\% in recall. Given the result, proposed methods trained on our dataset can be successfully launched in current NAVER shopping adult recognition system.

% Because managing a diverse set of deep learning models consumes an enormous amount of resources, 
Prior to employing \algname{}, deep learning models in NAVER Shopping used different models and data per task. Even for the same task, it is often necessary to train the image and text models separately and retrain them regularly whenever product distributions change over time, thus consuming enormous resources. Our proposed \algname{} framework can efficiently and jointly learn image and text representations through a single training process while showing comparable performance to previous models. 
% through zero-shot or only linear probing in most shopping tasks. 
Therefore, the proposed framework helps save vast resources by unifying NAVER Shopping's model lifecycle from model management, training, validation, and serving.

% managing fewer models, higher accuracy via multi-task learning, and usage as a transferable backbone model for a new downstream task

% an enormous set of resources are used 
% using our models 
 
% from a day to at most a year.

% \subsection{Ablation tests} 
% (hard negative 잘나오면)
%     - soft label vs hard label
%     - hard negative

% \begin{table*}
% \begin{minipage}{1.3\columnwidth}
    
% \end{minipage}\hfill % maximize the horizontal separation
% \begin{minipage}{0.7\columnwidth}
    
%   \end{minipage}
% \end{table*}

%% file: tab-model.tex
{
\newcolumntype{L}[1]{>{\raggedright\let\newline\\\arraybackslash\hspace{0pt}}m{#1}}
\newcolumntype{X}[1]{>{\centering\let\newline\\\arraybackslash\hspace{0pt}}p{#1}}
\begin{table}[!t]
\centering
\resizebox{\linewidth}{!}{%

\begin{tabular}{l|cc|cc|c} \toprule
\multirow{2}{*}{Method} & \multicolumn{2}{c|}{Text model} & \multicolumn{2}{c|}{Image model} & \# Pairs \\
 & Hidden dim & \# Params & Model & \# Params &  \\ \midrule
BERT-multi & 768 & 167M & - & - & - \\
BERT-dlarva & 768 & 110M & - & - & - \\
CLIP & 512 & 37M & ViT/B-32 & 87M & 400M \\
KELIP & 512 & 88M & ViT/B-32 & 87M & 1.1B \\
e-LiT & 768 & 167M & ViT/B-32 & 87M & 270M \\
e-CLIP (filter) & 768 & 167M & ViT/B-32 & 87M & 270M \\
e-CLIP  & 768 & 167M & ViT/B-32 & 87M & 330M \\
\bottomrule
\end{tabular}

}
\caption{Details of models and data size (\# pairs).}
\label{tab:model}
\vspace*{-0.7cm}
\end{table}
}

%  \hline

% \multirow{2}{*}{} & \multirow{2}{*}{\# Pairs in training set} & \multicolumn{3}{c}{Text model} & \multicolumn{2}{c}{Image model} \\ 
%  &  & \# Layers & Hidden dim & \# Params & Model & \# Params \\ \toprule
% BERT-multi & - & 12 & 768 & 167,356,416 & - & - \\
% BERT-dlarva & - & 12 & 768 & 110,620,416 & - & - \\
% CLIP & 400M & 12 & 512 & 37,828,608 & ViT/B-32 & 87,849,216 \\
% KELIP & 1.1B & 12 & 512 & 88,463,872 & ViT/B-32 & 87,849,216 \\
% e-LiT & 270M & 12 & 768 & 167,356,416 & ViT/B-32 & 87,849,216 \\
% \algname{} (hard) & 270M & 12 & 768 & 167,356,416 & ViT/B-32 & 87,849,216 \\
% \algname{} (soft) & 330M & 12 & 768 & 167,356,416 & ViT/B-32 & 87,849,216 \\

%% file: tab-catalog-match.tex
{
\newcolumntype{L}[1]{>{\raggedright\let\newline\\\arraybackslash\hspace{0pt}}m{#1}}
\newcolumntype{X}[1]{>{\centering\let\newline\\\arraybackslash\hspace{0pt}}p{#1}}

\begin{table}[t]
\centering
\resizebox{\linewidth}{!}{%

\begin{tabular}{l|ccc|ccc} \toprule
Category & \multicolumn{3}{c|}{Electronics} & \multicolumn{3}{c}{Food} \\ 
Modality & Text & Image & \begin{tabular}[x]{@{}c@{}}Multi-\\modal\end{tabular} & Text & Image & \begin{tabular}[x]{@{}c@{}}Multi-\\modal\end{tabular} \\ \toprule
% BERT-larva & 13.695 & - & - & 19.941 & - & - \\
BERT-multi & 17.4 & - & - & 24.3 & - & - \\
BERT-larva & 14.0 & - & - & 23.7 & - & - \\
CLIP & 30.9 & 36.5 & 40.2 & 30.8 & 41.9 & 43.0 \\
KELIP & 33.5 & 35.6 & 41.2 & 53.2 & 38.7 & 52.9 \\
e-LiT & 41.1 & \textbf{36.6} & 46.4 & 66.7 & 42.7 & 66.8 \\
\algname{} (filter) & 39.6 & 32.9 & 43.7 & 64.9 & 49.4 & 66.1 \\
\algname{} & 43.4 & 33.2 & 46.5 & 67.0 & 48.6 & 67.3 \\
\algname{} (hard) & \textbf{46.6} & 34.5 & \textbf{48.7} & \textbf{68.2} & \textbf{50.9} & \textbf{68.9} \\
\bottomrule
\end{tabular}

}
\caption{Top-1 accuracy\,(\%) of product matching.}
\vspace*{-0.9cm}
\label{tab:catalog-match}
\end{table}
}

%% file: tab-catalog-cluster.tex
{
\newcolumntype{L}[1]{>{\raggedright\let\newline\\\arraybackslash\hspace{0pt}}m{#1}}
\newcolumntype{X}[1]{>{\centering\let\newline\\\arraybackslash\hspace{0pt}}p{#1}}
\begin{table*}[t]
\centering
% \resizebox{\linewidth}{!}{%

\begin{tabular}{l|l|ccc|ccc|ccc} \toprule
\multicolumn{2}{l}{Modality} & \multicolumn{3}{c}{Text} & \multicolumn{3}{c}{Image} & \multicolumn{3}{c}{Multimodal} \\
\multicolumn{2}{l}{Metric} & ACC\,($\uparrow$) & NMI\,($\uparrow$) & ARI\,($\uparrow$) & ACC\,($\uparrow$) & NMI\,($\uparrow$) & ARI\,($\uparrow$) & ACC\,($\uparrow$) & NMI\,($\uparrow$) & ARI\,($\uparrow$) \\
 \toprule
\multirow{8}{*}{Electronics} & BERT-multi & 0.30113 & 0.69736 & 0.15817 & - & - & - & - & - & - \\
 & BERT-larva & 0.27077 & 0.69358 & 0.14677 & - & - & - & - & - & - \\
& CLIP & 0.52982 & 0.86198 & 0.43697 & 0.52150 & 0.85948 & 0.41423 & 0.58190 & 0.88571 & 0.49172 \\
& KELIP & 0.56415 & 0.88767 & 0.50492 & 0.52818 & 0.86788 & 0.43175 & 0.59765 & 0.90290 & 0.53277 \\
& e-LiT & 0.61519 & 0.91088 & 0.57202 & 0.51479 & 0.85446 & 0.40188 & 0.58451 & 0.90305 & 0.52882 \\
& \algname{} (filter) & 0.59840 & 0.90691 & 0.55877 & 0.54582 & 0.87710 & 0.45887 & 0.57940 & 0.90275 & 0.53720 \\
& \algname{} & 0.60619 & 0.91398 & 0.55831 & 0.52789 & 0.87488 & 0.44171 & 0.58197 & 0.90831 & 0.53834 \\
& \algname{} (hard) & \textbf{0.63230} & \textbf{0.92481} & \textbf{0.60243} & \textbf{0.55043} & \textbf{0.88116} & \textbf{0.46199} & \textbf{0.61579} & \textbf{0.91609} & \textbf{0.56664} \\
\bottomrule

\multirow{8}{*}{Food} & BERT-multi & 0.32305 & 0.72799 & 0.17194 & - & - & - & - & - & - \\
& BERT-larva & 0.32978 & 0.74063 & 0.18663 & - & - & - & - & - & - \\
& CLIP & 0.54439 & 0.87683 & 0.44835 & 0.54919 & 0.87505 & 0.44374 & 0.54076 & 0.87343 & 0.43057  \\
& KELIP & 0.63662 & 0.92114 & 0.59224 & 0.49241 & 0.84651 & 0.37045 & 0.59227 & 0.90459 & 0.53638 \\
& e-LiT & 0.70263 & 0.94864 & 0.68430 & 0.54838 & 0.87436 & 0.43045 & 0.66373 & 0.93910 & 0.63156 \\
& \algname{} (filter) & \textbf{0.71346} & 0.95205 & \textbf{0.70110} & 0.62218 & 0.92105 & 0.56723 & 0.68449 & 0.94837 & 0.65927 \\
& \algname{} & 0.69136 & 0.94825 & 0.66954 & 0.62607 & 0.92190 & 0.56503 & 0.71015 & 0.95184 & 0.68574 \\
& \algname{} (hard) & 0.71091 & \textbf{0.95235} & 0.69348 & \textbf{0.63204} & \textbf{0.92443} & \textbf{0.58422} & \textbf{0.72024} & \textbf{0.95440} & \textbf{0.69640} \\
\bottomrule
\end{tabular}

% }
\caption{Test performance of product clustering with respect to ACC, NMI, and ARI. .}
\label{tab:catalog-cluster}
\vspace*{-0.65cm} %%% \vspace*{-0.75cm}
\end{table*}
}

%% file: tab-attribute.tex
{
\newcolumntype{L}[1]{>{\raggedright\let\newline\\\arraybackslash\hspace{0pt}}m{#1}}
\newcolumntype{X}[1]{>{\centering\let\newline\\\arraybackslash\hspace{0pt}}p{#1}}
\begin{table}[!t]
\centering
\resizebox{\linewidth}{!}{%
\begin{tabular}{l|ccccc} \toprule
Attribute  & Colors & Materials & \begin{tabular}[x]{@{}c@{}}Heel\\types\end{tabular} & \begin{tabular}[x]{@{}c@{}}Toe\\shapes\end{tabular} & Patterns  \\ \toprule
CLIP & 70.2 & 46.1 & 42.0 & 41.8 & 90.2 \\
KELIP & 80.2 & 74.4 & 71.0 & 80.7 & \textbf{92.9}  \\
e-LiT & 77.7 & 66.6 & 59.7 & 57.6 & 91.2 \\
\algname{} (filter) & \textbf{80.5} & 77.5 & 77.8 & 82.3 & 90.9  \\
\algname{}  & 78.5 & 72.7 & 77.8 & 82.1 & 92.4 \\
\algname{} (hard) & 79.5 & \textbf{77.8} & \textbf{78.6} & \textbf{82.4} & \textbf{92.9} \\
\bottomrule
\end{tabular}
}
\caption{Test accuracy\,(\%) of zero-shot attribute extraction. }
\label{tab:attribute}
\vspace*{-0.8cm} %%% \vspace*{-0.9cm}
\end{table}
}

%% file: tab-category-zero.tex
{
\newcolumntype{L}[1]{>{\raggedright\let\newline\\\arraybackslash\hspace{0pt}}m{#1}}
\newcolumntype{X}[1]{>{\centering\let\newline\\\arraybackslash\hspace{0pt}}p{#1}}
\begin{table*}[!t]
\centering
% \resizebox{\linewidth}{!}{%
\begin{tabular}{l|X{1.0cm}X{1.2cm}c|X{1.2cm}X{1.2cm}c|X{1.2cm}X{1.2cm}c}  \toprule
Category & \multicolumn{3}{c}{All} & \multicolumn{3}{c}{Fashion} & \multicolumn{3}{c}{Children} \\ \cline{2-10} 
Modality & Text & Image & Multimodal & Text & Image & Multimodal & Text & Image & Multimodal \\ \toprule
CLIP & 0.5 & 0.5 & 0.6 & 0.7 & 0.0 & 0.5 & 0.5 & 0.0 & 0.4 \\
KELIP & 28.9 & 17.9 & 28.7 & 42.6 & 29.7 & 43.0 & 36.0 & 24.9 & 36.2 \\
e-LiT & 20.9 & 31.9 & 35.2 & 34.4 & 37.7 & 42.5 & 32.0 & 31.5 & 38.8 \\
\algname{} (filter) & 30.2 & \textbf{37.0} & \textbf{40.5} & \textbf{62.0} & \textbf{59.8} & \textbf{67.9} & \textbf{51.6} & \textbf{48.6} & \textbf{56.4} \\
\algname{} & \textbf{32.0} & 32.7 & 36.3 & 45.4 & 48.7 & 53.3 & 39.0 & 39.7 & 45.2 \\
\algname{} (hard)  & 23.3 & 24.5 & 27.0 & 50.9 & 47.8 & 56.8 & 40.4 & 44.9 & 37.4 \\
\bottomrule
\end{tabular}

% }
\caption{Test classification accuracy\,(\%) of zero-shot transfer for category classification.}
\label{tab:category-zero}
\vspace*{-0.65cm}
\end{table*}
}

% \begin{comment}
% {
% \newcolumntype{L}[1]{>{\raggedright\let\newline\\\arraybackslash\hspace{0pt}}m{#1}}
% \newcolumntype{X}[1]{>{\centering\let\newline\\\arraybackslash\hspace{0pt}}p{#1}}
% \begin{table*}[!t]
% \centering
% % \resizebox{\linewidth}{!}{%
% \begin{tabular}{l|ccc|ccc|ccc}  \toprule
% Category & \multicolumn{3}{c}{All} & \multicolumn{3}{c}{Fashion} & \multicolumn{3}{c}{Children} \\ \cline{2-10} 
% Modality & Text & Image & Multimodal & Text & Image & Multimodal & Text & Image & Multimodal \\ \toprule
% CLIP & 0.483 & 0.500 & 0.565 & 0.649 & 0.016 & 0.514 & 0.534 & 0.0213 & 0.436 \\
% KELIP & 28.943 & 17.928 & 28.706 & 42.560 & 29.699 & 43.056 & 35.954 & 24.944 & 36.227 \\
% e-LiT & 20.909 & 31.884 & 35.168 & 34.418 & 37.737 & 42.464 & 32.000 & 31.496 & 38.783 \\
% \algname{} (filter) & 30.229 & \textbf{36.962} & \textbf{40.459} & \textbf{62.009} & \textbf{59.805} & \textbf{67.904} & \textbf{51.553} & \textbf{48.643} & \textbf{56.420} \\
% \algname{} & \textbf{31.982} & 32.670 & 36.323 & 45.449 & 48.705 & 53.300 & 38.965 & 39.659 & 45.237 \\
% \algname{} (hard)  & 23.342 & 24.505 & 26.994 & 50.869 & 47.803 & 56.840 & 40.437 & 44.923 & 37.356 \\
% \bottomrule
% \end{tabular}

% % }
% \caption{Test accuracy\,(\%) of zero-shot transfer for category classification.}
% \label{tab:category-zero}
% \vspace*{-0.65cm}
% \end{table*}
% }

% \end{comment}

%% file: tab-category-linear.tex
{
\newcolumntype{L}[1]{>{\raggedright\let\newline\\\arraybackslash\hspace{0pt}}m{#1}}
\newcolumntype{X}[1]{>{\centering\let\newline\\\arraybackslash\hspace{0pt}}p{#1}}
\begin{table*}[!t]
\centering
\resizebox{\linewidth}{!}{%
\begin{tabular}{l|ccc|ccc|ccc|ccc}
\toprule
Eval. Scheme & \multicolumn{6}{c|}{Linear probe} & \multicolumn{6}{c}{Finetuning} \\ \cline{2-13} 
Modality & \multicolumn{3}{c|}{Text} & \multicolumn{3}{c|}{Image} & \multicolumn{3}{c|}{Text} & \multicolumn{3}{c}{Image} \\ \cline{2-13}
Category & All & Fashion & Children & All & Fashion & Children & All & Fashion & Children & All & Fashion & Children \\ \toprule
BERT-multi & 44.9 & 60.5 & 52.4 & - & - & - & 60.3 & 81.2 & 71.2 & - & - & - \\
BERT-larva & 52.9 & 66.4 & 61.5 & - & - & - & 63.0 & 81.8 & 74.0 & - & - & - \\
CLIP & 0.1 & 0.1 & 0.1 & 45.0 & 60.6 & 50.6 & 0.2 & 0.0 & 0.1 & 50.8 & 71.1 & 61.8 \\
KELIP & 67.8 & 83.3 & 75.1 & 55.4 & 73.7 & 61.9 & 67.4 & 83.3 & 76.1 & 53.6 & 74.2 & 64.5 \\
e-LiT & 69.4 & 83.6 & 77.0 & 44.0  & 50.8  & 59.3 & \textbf{70.0} & \textbf{84.9} & 78.2 & 49.8  & 60.4 & 70.3 \\
\algname{} (filter) & 69.5 & 84.3 & 77.2 & \textbf{60.9} & 77.0 & 66.5 & 68.9 & 84.5 & 78.0 & 57.8 & 76.4 & 67.6 \\
\algname{} & \textbf{70.3} & \textbf{85.0} & \textbf{77.5} & 60.6 & \textbf{77.1} & \textbf{66.8} & 69.1 & 84.7 & \textbf{78.3} & \textbf{58.6} & \textbf{76.7} & \textbf{67.8} \\
\algname{} (hard) & 69.6 & 84.7 & 77.1 & 60.5 & 76.6 & 66.5 & 69.1 & 84.7 & 78.2 & 58.1 & 76.4 & 67.8 \\
\bottomrule
\end{tabular}
}

\caption{Test classification accuracy\,(\%) of linear probe and fine-tuning for category classification.}
\label{tab:category-linear}
\vspace*{-0.65cm}
\end{table*}
}

%% file: tab-adult-linear.tex
{
\newcolumntype{L}[1]{>{\raggedright\let\newline\\\arraybackslash\hspace{0pt}}m{#1}}
\newcolumntype{X}[1]{>{\centering\let\newline\\\arraybackslash\hspace{0pt}}p{#1}}
\begin{table}[!t]
\centering
% \resizebox{\linewidth}{!}{%
\begin{tabular}{l|X{1.7cm}|X{1.7cm}|X{1.7cm}} \toprule 
Eval. Scheme & Zero-shot & Linear Probe & Finetuning \\ 
 \toprule
CLIP & 51.5 & 71.2 & 79.8 \\
KELIP & 57.5 & 73.0 & 81.3 \\
e-Lit & 53.2 & 71.2 & 79.8 \\
\algname{} (filter) & 61.1 & 76.6 & 84.9 \\
\algname{}  & \textbf{63.8} & 78.4 & 83.8 \\
\algname{} (hard)&63.7&\textbf{78.4}&\textbf{85.1}\\
\bottomrule
\end{tabular}
% }
\caption{F1-scores for adult product recognition.}
\label{tab:adult-linear}
\vspace*{-0.9cm}
\end{table}
}

%% file: 6-conclusion.tex
% \vspace*{-0.3cm}

\section{Conclusion}

% 9pg (반장)
% Conclusion + Future work
% In this paper, we have presented approaches to train a unified multimodal model by jointly training image and text models. 

We proposed \algname{} to train a unified multimodal model by jointly training image and text models for large-scale e-commerce systems. We shared innovative ideas for preparing large training data through preprocessing and how to accelerate training and convergence with high accuracy. 
We evaluated the learned representations by transferring the model to important downstream tasks in the industry and demonstrate the effectiveness of \algname{} with significant improvement over other competitive pre-training methods.
 
%  . The results indicate that the proposed techniques can enhance performance and applied to tasks in the industry. 